\documentclass[conference]{IEEEtran}
\usepackage{graphicx}
\usepackage{algorithm}
\usepackage{algpseudocode}
\usepackage[tight,footnotesize,nooneline]{subfigure}
\usepackage{url}
\usepackage{ctable}

\begin{document}
%
\title{Evolution of Ideas: A Novel Memetic Algorithm Based on Semantic Networks}


%

\author{\IEEEauthorblockN{At{\i}l{\i}m~G\"{u}ne\c{s}~Baydin\IEEEauthorrefmark{1}\IEEEauthorrefmark{2}}\IEEEauthorblockA{\IEEEauthorrefmark{2}Departament d'Enginyeria de la Informaci\'{o} i de les Comunicacions\hspace{4mm}\\Universitat Aut\`{o}noma de Barcelona\\ 08193 Bellaterra, Spain\\Email: gunesbaydin@iiia.csic.es}\and\IEEEauthorblockN{Ramon~L\'{o}pez~de~M\'{a}ntaras\IEEEauthorrefmark{1}}\IEEEauthorblockA{\IEEEauthorrefmark{1}Artificial Intelligence Research Institute, IIIA\,-\,CSIC\\Campus Universitat Aut\`{o}noma de Barcelona\\08193 Bellaterra, Spain\\Email: mantaras@iiia.csic.es}}

\maketitle

\begin{abstract}
This paper presents a new type of evolutionary algorithm (EA) based on the concept of ``meme'', where the individuals forming the population are represented by semantic networks and the fitness measure is defined as a function of the represented knowledge. Our work can be classified as a novel memetic algorithm (MA), given that (1) it is the units of culture, or information, that are undergoing variation, transmission, and selection, very close to the original sense of memetics as it was introduced by Dawkins; and (2) this is different from existing MA, where the idea of memetics has been utilized as a means of local refinement by individual learning after classical global sampling of EA. The individual pieces of information are represented as simple semantic networks that are directed graphs of concepts and binary relations, going through variation by memetic versions of operators such as crossover and mutation, which utilize knowledge from commonsense knowledge bases. In evaluating this introductory work, as an interesting fitness measure, we focus on using the structure mapping theory of analogical reasoning from psychology to evolve pieces of information that are analogous to a given base information. Considering other possible fitness measures, the proposed representation and algorithm can serve as a computational tool for modeling memetic theories of knowledge, such as evolutionary epistemology and cultural selection theory.
\end{abstract}

\section{Introduction}
  The idea that a simple progression of variation, natural selection, and heredity can account for the great complexity and apparent design observed in living beings has eventually led to the formulation of \emph{Universal Darwinism}, generalizing the mechanisms and extending the domain of this process to systems outside biology, including economics, psychology, physics, and even culture \cite{Dennett1995,Bickhard2003}. Within this larger framework, the concept of \emph{meme} introduced by Dawkins as an evolving unit of culture---or information, idea, or belief---analogous to a \emph{gene} \cite{Dawkins1989}, hosted, altered, and reproduced in individuals' minds, forms the basis of the field of memetics\footnote{Quoting Dawkins \cite{Dawkins1989}: \emph{``Examples of memes are tunes, ideas, catch-phrases, clothes fashions, ways of making pots or of building arches. Just as genes propagate themselves in the gene pool by leaping from body to body via sperms or eggs, so memes propagate themselves in the meme pool by leaping from brain to brain...''}}.

  Within the discipline of evolutionary computation, the recently maturing field of memetic algorithms (MA) has experienced increasing interest as a successful method for solving many hard optimization problems \cite{Moscato1989,Moscato2004,Krasnogor2005}. The existing formulation of MA is essentially a hybrid approach, combining classical evolutionary algorithms (EA) with local search, where the population-based global sampling of EA in each generation is followed by a local search, or learning, performed by each candidate solution. For this reason, this approach has been often referred to under different names besides MA, such as ``hybrid EAs'' or ``Lamarckian EAs''. To date, MAs have been successfully applied to a wide variety of problem domains such as NP-hard optimization problems \cite{Bui1996,Merz2002}, engineering \cite{Cotta2001}, machine learning \cite{Abbass2001,Mignotte2000}, and robotics \cite{Chaiyaratana1999}.

\begin{figure}[!t]
    \centering
    \includegraphics[width=3.2in]{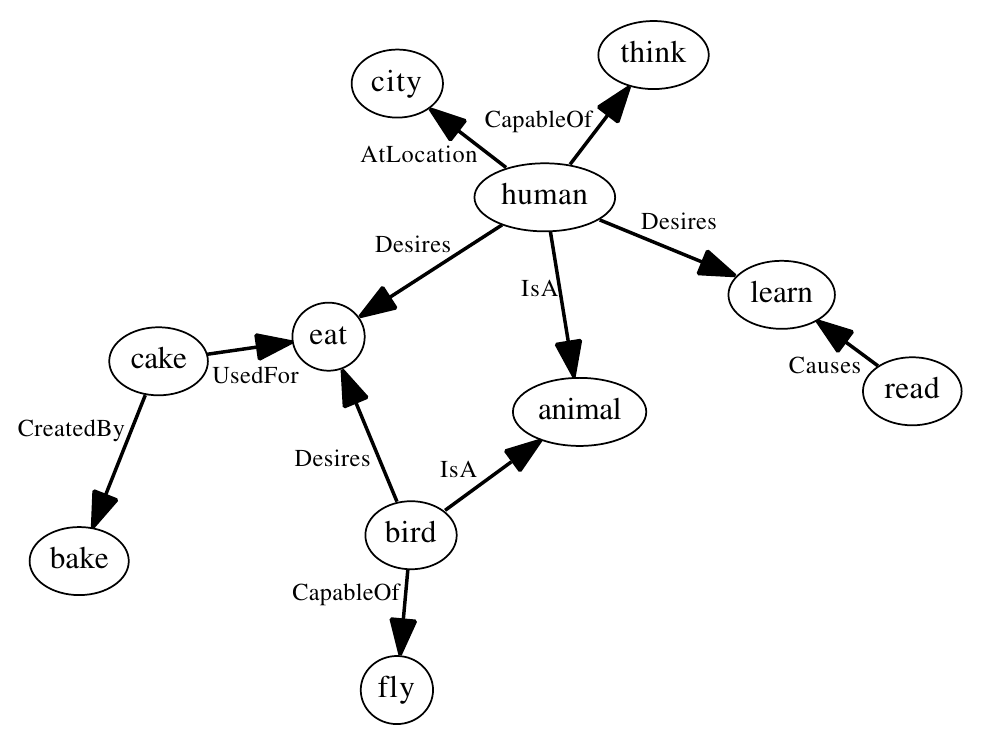}
    \caption{A semantic network with 11 concepts and 11 relations.}
    \label{FigureSemanticNetwork}
  \end{figure}

  The aim of this study is to propose a computational model comprising a \emph{meme pool} subject to variation and selection that will be able to evolve pieces of knowledge under a given memetic fitness measure, paralleling the existing use of EA in solving optimization problems. While this approach is based on memetics, it is unlike the existing sense of the word in current MA (as an hybridization of local search into EA). Rather, it is intended as a new tool focused exclusively on the memetic evolution of knowledge itself, which can find use in fields such as knowledge-based systems, reasoning, and computational creativity. As the basis of our approach, we introduce a solution representation based on semantic networks \cite{Sowa1991}. These are a simple type of formal representation for ontologies, formed by graphs where vertices correspond to concepts and edges correspond to directed relations (Fig.~\ref{FigureSemanticNetwork}). To operate on these structures, we adapt variation operators such as crossover and mutation to manipulate concepts and relations, utilizing the commonsense knowledge bases of ConceptNet \cite{Havasi2007} and WordNet \cite{Fellbaum1998}. From the perspective of representation and genetic operators, our approach is partly similar to the use of tree structures in genetic programming \cite{Koza2003} and the approaches of genetic network programming \cite{Katagiri2002,Mabu2007}, parallel distributed genetic programming \cite{Poli1999}, and evolutionary graph generation \cite{Chen2002}.

  In this paper, we present three main contributions:
  \begin{itemize}
  \item Using semantic networks for encoding individual \emph{memotypes}, where graphs formed by concepts and relations represent units of evolving knowledge.
  \item Introducing operations for evolutionary variation, such as mutation and crossover, that are adapted to work on semantic networks.
  \item Introducing a memetic fitness measure for evolutionary selection to evaluate our implementation, based on the \emph{structure mapping theory} from psychology.
  \end{itemize}

  Following an overview of our approach, the paper presents implementation details of the proposed algorithm in Section~\ref{SectionNewAlgorithm}. We evaluate the method with experiments presented in Section~\ref{SectionExperimentalResults} and conclude the paper and discuss future work in Section~\ref{SectionConclusion}.

\section{A New Type of Memetic Algorithm}
  \label{SectionNewAlgorithm}

  Our algorithm proceeds similar to conventional EA, with a relatively small set of parameters. We implement semantic networks as linked-list data structures of concept and relation objects. The descriptions of representation, memetic variation, fitness evaluation, and selection steps are presented in the following sections. The parameters affecting each step of the algorithm (Algorithm~\ref{AlgorithmNew}) are given in Table~\ref{TableParameters}.

  \begin{algorithm}
  \caption{The proposed memetic algorithm}
  \label{AlgorithmNew}
  \begin{algorithmic}[1]
  \Procedure{MemeticAlgorithm}{}
    \State $P(t=0) \gets$ \Call{Initialize}{$Pop_{size}, C_{max}, R_{min}, T$}
    \Repeat
      \State $\phi(t) \gets$ \Call{EvaluateFitnesses}{$P(t)$}
      \State $S(t) \gets$ \Call{Selection}{$P(t), \phi(t), S_{size}, S_{prob}$}
      \State $V(t) \gets$ \Call{Variation}{$S(t), P_c, P_m, T$}
      \State $P(t+1) \gets V(t)$
      \State $t \gets t+1$
    \Until{stop criterion}
  \EndProcedure
  \end{algorithmic}
  \end{algorithm}

  \subsection{Representation: Semantic Networks as Memes}
  \label{SectionRepresentation}
  The algorithm is centered on the use of semantic networks \cite{Sowa1991} for encoding evolving memotypes. A semantic network is a graphic notation for the representation of knowledge in the form of sets of vertices representing concepts, interconnected by edges representing relations (Fig.~\ref{FigureSemanticNetwork}). This type of graph representation has found use in many subfields of artificial intelligence, including natural language processing, machine translation, and information retrieval. Constructs resembling semantic networks have long been in use also in other fields such as philosophy and linguistics.

  An important characteristic of a semantic network is whether it is definitional or assertional: in definitional networks the emphasis is on taxonomic relations (e.g. $IsA(bird, animal)$\footnote{Here we adopt the notation $IsA(bird, animal)$ to mean that the concepts $bird$ and $animal$ are connected by the directed relation $IsA$, i.e. ``bird is an animal.''}) describing a subsumption hierarchy that is true by definition; in assertional networks, the relations describe instantiations and assertions that are contingently true (e.g. $AtLocation(human, city)$) \cite{Sowa1991}. In this study we combine the two approaches for increased expressivity (Fig.~\ref{FigureSemanticNetwork}). As such, semantic networks provide a simple yet powerful means to represent the ``memes'' of Dawkins as data structures that are algorithmically manipulatable, allowing a procedural implementation of memetic evolution.

  There are several existing algorithms using graph-based representations for the encoding of candidate solutions in EA \cite{Montes2004}. The most notable work among these is genetic programming (GP) \cite{Koza2003}, where candidate solutions are pieces of computer program represented in a tree hierarchy, which is actually a specific type of graph structure \cite{Montes2004}. In parallel distributed genetic programming (PDGP) \cite{Poli1999}, the restrictions of the tree structure of GP is relaxed by allowing multiple outputs from a node, which allows a high degree of parallelism in the evolved programs. In evolutionary graph generation (EGG) \cite{Chen2002} the focus is on evolving graphs with applications in electronic circuit design. Genetic network programming (GNP) \cite{Katagiri2002,Mabu2007} introduces compact networks with conditional branching and action nodes; and similarly, neural programming (NP) \cite{Teller1998} combines GP with artificial neural networks for the discovery of network structures via evolution.

  The use of a graph-based representation makes the design of variation operators specific to graphs necessary. In works such as GNP, this is facilitated by using a string-based encoding of node names, types, and connectivity, permitting operators very close to their counterparts in conventional EA; and in PDGP, the operations are simplified by making nodes occupy points in a fixed-size two-dimensional grid. What is common with GP related algorithms is that the output of each node in the graph can constitute an input to another node. In comparison, the range of connections that can form a semantic network of a given set of concepts is limited by commonsense knowledge, i.e. the relations have to make sense to be useful (e.g. $IsA(bird, animal)$ is meaningful while $Causes(bird, table)$ is not). To address this issue, we introduce new crossover and mutation operations for memetic variation, making use of commonsense reasoning \cite{Mueller2006,Havasi2007} and adapted to work on semantic networks.

  \subsection{Commonsense Knowledge Bases}

  Commonsense reasoning refers to the type of reasoning involved in everyday human thinking, based on \emph{commonsense knowledge} that an ordinary person is expected to know, or ``the knowledge of how the world works'' \cite{Mueller2006}. Commonsense knowledge bases such as the \emph{ConceptNet}\footnote{\url{http://conceptnet.media.mit.edu}} project of MIT Media Lab \cite{Havasi2007} and \emph{Cyc}\footnote{\url{http://www.cyc.com}} maintained by Cycorp company are set up to assemble and classify commonsense information. The lexical database \emph{WordNet}\footnote{\url{http://wordnet.princeton.edu}} maintained by the Cognitive Science Laboratory at Princeton University also has characteristics of a commonsense knowledge base, via synonym, hypernym\footnote{Y is a \emph{hypernym} of X if every X is a (kind of) Y ($IsA(dog, canine)$).}, and hyponym\footnote{Y is a \emph{hyponym} of X if every Y is a (kind of) X.} relations \cite{Fellbaum1998}.

  In our implementation we make use of ConceptNet version 4 and WordNet version 3 to obtain and process commonsense knowledge, where ConceptNet contributes around 560,000 definitional and assertional relations involving 320,000 concepts and WordNet contributes definitional relations involving around 117,000 synsets\footnote{A \emph{synset} is a set of synonyms that are interchangeable without changing the truth value of any propositions in which they are embedded.}, as of the writing of this article. The hypernym and hyponym relations among noun synsets in WordNet provide a reliable collection of $IsA$ relations. In contrast, the variety of assertions in ConceptNet, contributed by volunteers across the world, makes it more prone to noise. We address this by ignoring all assertions with a reliability score (determined by contributors' voting) below a set minimum $R_{min}$ (Table~\ref{TableParameters}).

  \subsection{Initialization}
  \label{SectionInitialization}

  At the start of a run, the population of size $Pop_{size}$ is initialized with individuals created by \emph{random semantic network generation} (Algorithm~\ref{AlgorithmNew}). This is achieved by starting from a network comprising only one concept randomly picked from commonsense knowledge bases and running a semantic network expansion algorithm that (1) randomly picks a concept in the given network (e.g. $human$); (2) compiles a list of relations---from commonsense knowledge bases---that the picked concept can be involved in (e.g. $\{CapableOf(human, think), Desires(human, eat), \cdots\}$) (3) appends to the network a relation randomly picked from this list, together with the other involved concept; and (4) repeats this until a given number of concepts has been appended or a set timeout $T$ has been reached (covering situations where there are not enough relations). Figure~\ref{FigureSemanticNetwork} presents a random semantic network created this way. Note that even if grown in a random manner, the network itself is totally meaningful because it is a combination of information from commonsense knowledge bases.

  The initialization algorithm depends upon the parameters of $C_{max}$, the maximum number of initial concepts, and $R_{min}$, the minimum ConceptNet relation score (Table~\ref{TableParameters}).

  \subsection{Selection}

  After the assignment of a fitness value to each individual in the current generation (Section~\ref{SectionFitness}), all individuals in the population are replaced with offspring generated by variation operators applied on parents. The parents are probabilistically selected from the population according to their fitness, with reselection allowed. While individuals with a higher fitness have a better chance of being selected, even individuals with low fitness have a chance to produce offspring, however small. In our experiments we employ tournament selection (Section~\ref{SectionExperimentalResults}).

  In each cycle of the algorithm, crossover is applied to parents selected from the population until $Pop_{size} \times P_c$ offspring are created (Table~\ref{TableParameters}). Mutation is applied to $Pop_{size} \times P_m$ selected individuals, supplying the remaining part of the next generation (i.e. $P_c + P_m = 1$). We also employ elitism, by replacing a randomly picked offspring in the next generation with the individual with the current best fitness.

  \subsection{Memetic Variation Operators}

  In contrast with existing graph-structured evolutionary approaches such as GP, PDGP, and GNP that we have discussed in Section~\ref{SectionRepresentation}, our representation does not permit arbitrary connections between different nodes and requires variation operators that should be based on information provided by commonsense knowledge bases.

  This means that any variation operation on the individuals should: (1) preserve the structure within boundaries set by commonsense knowledge; and (2) ensure that even vertices and edges randomly introduced into a semantic network connect to existing ones through meaningful relations.

  Here we present commonsense crossover and mutation operators specific to semantic networks.

  \subsubsection{Commonsense Crossover}

  In classical EA, features representing individuals are commonly encoded as linear strings and the crossover operation simulating genetic recombination is simply defined as a cutting and merging of this one dimensional object from two parents; and in graph-based approaches such as GP, subgraphs can be freely exchanged between parent graphs \cite{Pereira1999,Koza2003,Montes2004}. Here, as mentioned, the requirement that a semantic network has to make sense imposes significant constraints on the nature of recombination.

  We introduce two types of \emph{commonsense crossover} that are tried in sequence by the variation algorithm. The first type attempts a sub-graph interchange between two selected parents similar to common crossover in standard GP; and where this is not feasible due to the commonsense structure of relations forming the parents, the second type falls back to a combination of both parents into a new offspring.

\begin{figure*}[!t]
    \centering
    \subfigure[Parent 1]{\includegraphics[width=2.2in]{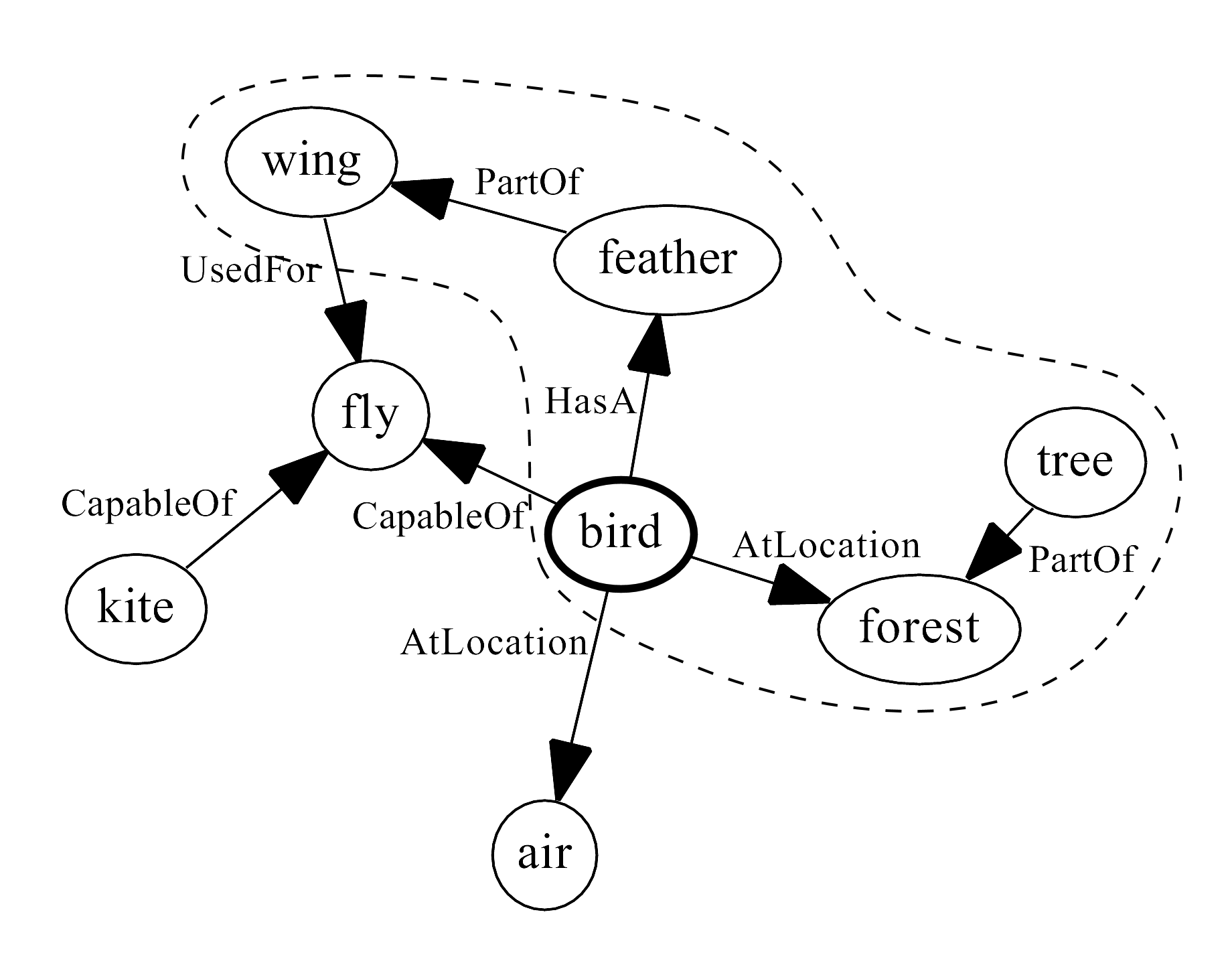}}
    \hfil
    \subfigure[Parent 2]{\includegraphics[width=2.2in]{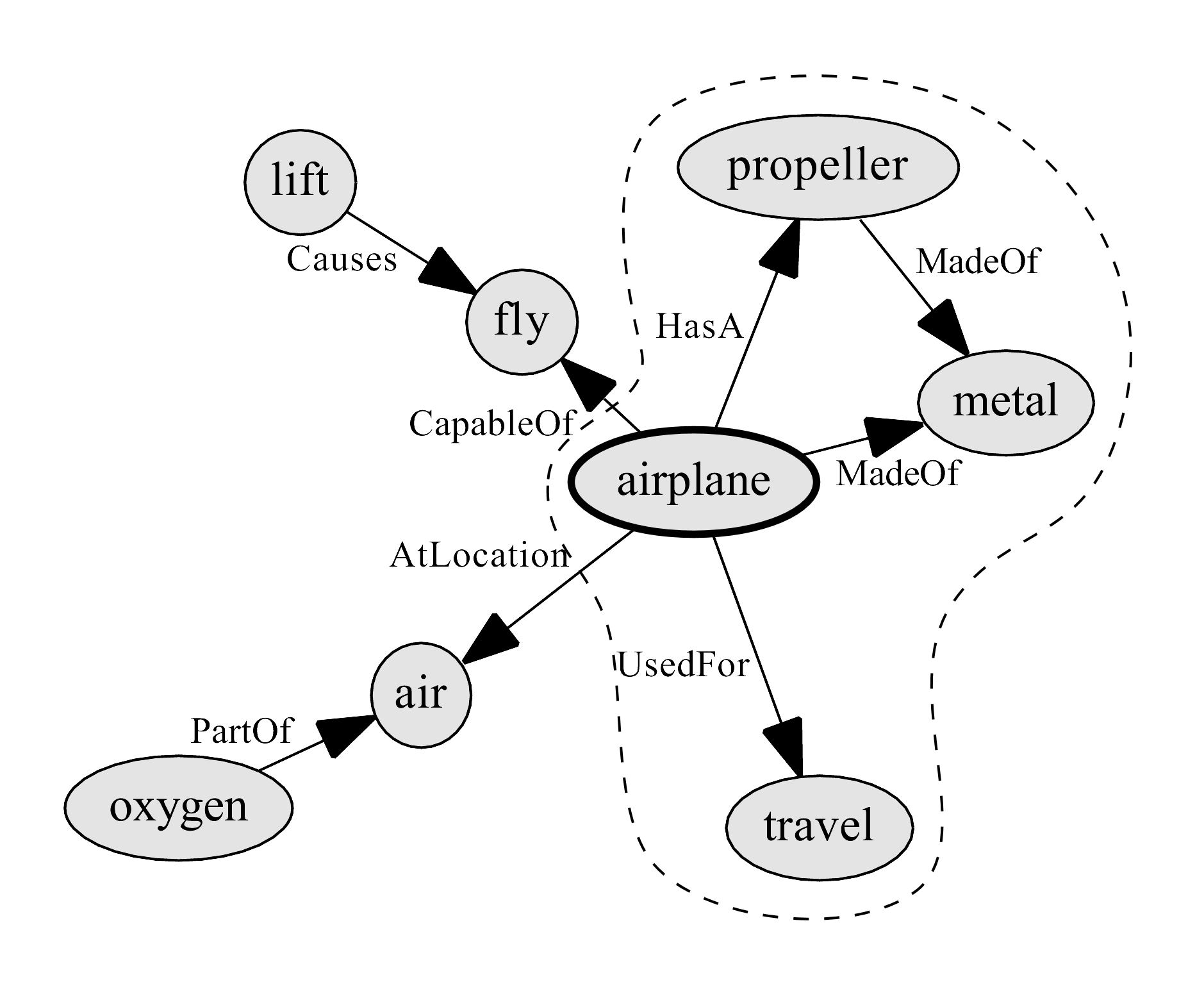}}\\
    \subfigure[Offspring 1]{\includegraphics[width=2in]{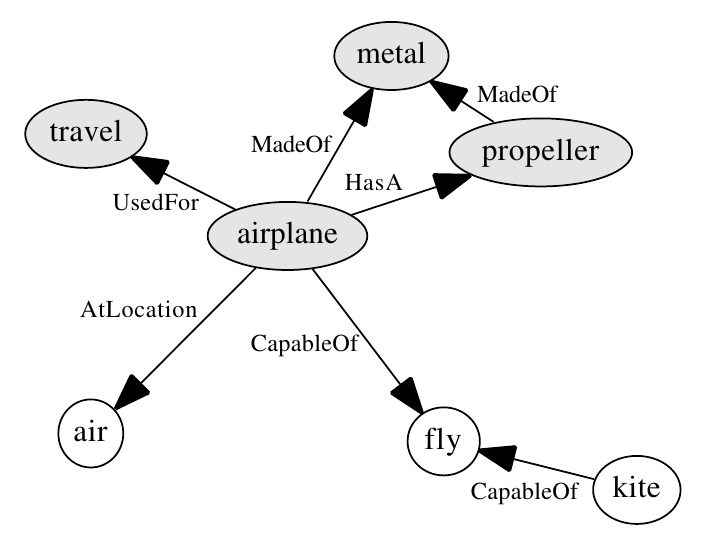}}
    \hfil
    \subfigure[Offspring 2]{\includegraphics[width=2.1in]{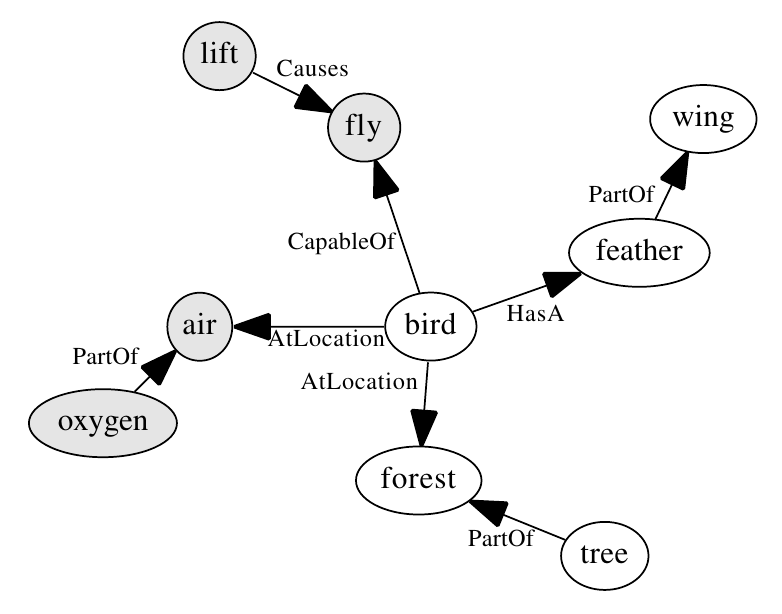}}
    \caption{Commonsense crossover type I (subgraph crossover), centered on the concepts of $bird$ for parent 1 and $airplane$ for parent 2.}
    \label{FigureCrossoverTypeI}
  \end{figure*}

  \emph{Type I (subgraph crossover):} A pair of concepts, one from each parent, that are \emph{interchangeable}\footnote{We define two concepts from different semantic networks as \emph{interchangeable} if both can replace the other in all, or part, of the relations the other is involved in, queried from commonsense knowledge bases.} are selected as \emph{crossover concepts}, picked randomly out of all possible such pairs. For instance, in Figure~\ref{FigureCrossoverTypeI}, $bird$ and $airplane$ are interchangeable, since they can replace each other in the relations $CapableOf(\cdot, fly)$ and $AtLocation(\cdot, air)$. In each parent, a subgraph is formed, containing: (1) the crossover concept; (2) the set of all relations, and associated concepts, that are not common with the other crossover concept (In Figure~\ref{FigureCrossoverTypeI} (a), $HasA(bird, feather)$ and $AtLocation(bird, forest)$; and in (b) $HasA(airplane, propeller)$, $MadeOf(airplane, metal)$, and $UsedFor(airplane, travel)$); and (3) the set of all relations and concepts connected to these (In Figure~\ref{FigureCrossoverTypeI} (a) $PartOf(feather, wing)$ and $PartOf(tree, forest)$; and in (b) $MadeOf(propeller, metal)$), excluding the ones that are also one of those common with the other crossover concept (the concept $fly$ in Figure~\ref{FigureCrossoverTypeI} (a), because of the relation $CapableOf(\cdot, fly)$). This, in effect, forms a subgraph of information specific to the crossover concept, which is insertable into the other parent. Any relations between the subgraph and the rest of the network not going through the crossover concept are severed (e.g. $UsedFor(wing, fly)$ in Figure~\ref{FigureCrossoverTypeI} (a)). The two offspring are formed by exchanging these subgraphs between the parent networks (Figure~\ref{FigureCrossoverTypeI} (c) and (d)).

\begin{figure*}[!t]
    \centering
    \subfigure[Parent 1]{\includegraphics[width=1.7in]{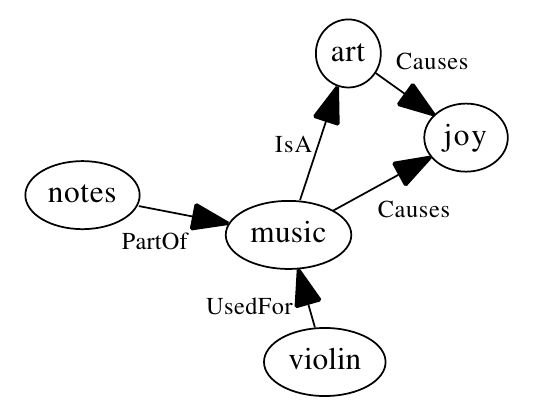}}
    \hfil
    \subfigure[Parent 2]{\includegraphics[width=1.6in]{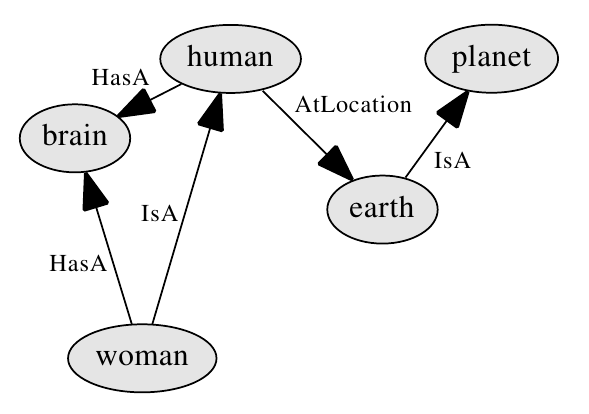}}
    \hfil
    \subfigure[Offspring]{\includegraphics[width=1.9in]{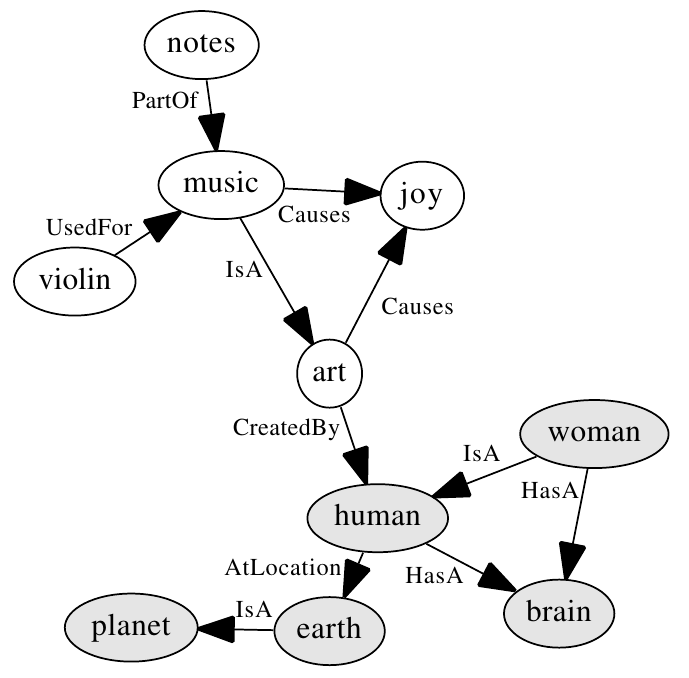}}
    \caption{Commonsense crossover type II (graph merging crossover), merging by the relation $CreatedBy(art, human)$. If no concepts attachable through commonsense relations are encountered, the offspring is formed by merging the parent networks as two separate clusters within the same semantic network.}
    \label{FigureCrossoverTypeII}
  \end{figure*}

  \emph{Type II (graph merging crossover):} A concept from each parent that is \emph{attachable}\footnote{We define a distinct concept as \emph{attachable} to a semantic network if at least one commonsense relation connecting the concept to any of the concepts in the network can be discovered from commonsense knowledge bases.} to the other parent is selected as a \emph{crossover concept}. The two parents are merged into an offspring by attaching a concept in one parent to another concept in the other parent, picked randomly out of all possible attachments ($CreatedBy(art, human)$ in Figure~\ref{FigureCrossoverTypeII}. Another possibility is $Desires(human, joy)$.). The second offspring is formed randomly the same way. In the case that no attachable concepts are found, the parents are merged as two separate clusters within the same semantic network. 

  \subsubsection{Commonsense Mutation}

  We introduce several types of \emph{commonsense mutation} operators that modify a parent by means of information from commonsense knowledge bases. For each mutation to be performed, the type is picked at random with uniform probability. If the selected type of mutation is not feasible due to the commonsense structure of the parent, another type is again picked. In the case that a set timeout of $T$ trials has been reached without any operation, the parent is returned as it is.

\begin{figure*}[!t]
    \centering
    \subfigure[Mutation type I (before)]{\parbox[t]{1.7in}{\centering\includegraphics[width=1.7in]{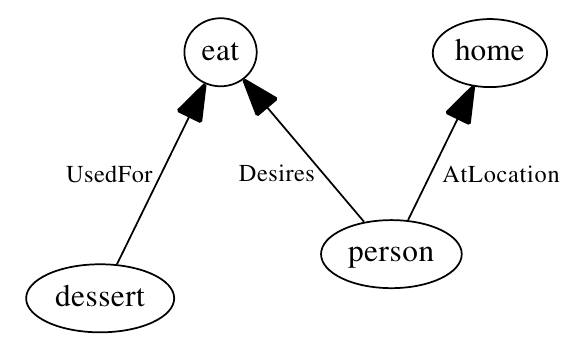}}}
    \hfil
    \subfigure[Mutation type I (after)]{\parbox[t]{1.7in}{\centering\includegraphics[width=1.7in]{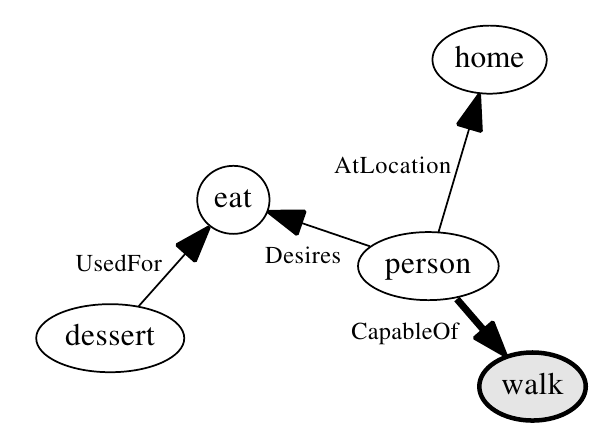}}}
    \hfil
    \subfigure[Mutation type IIa (before)]{\parbox[t]{1.7in}{\centering\includegraphics[width=1.3in]{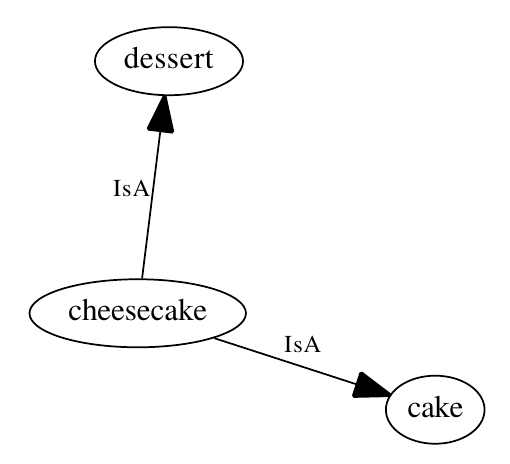}}}
    \hfil
    \subfigure[Mutation type IIa (after)]{\parbox[t]{1.7in}{\centering\includegraphics[width=1.3in]{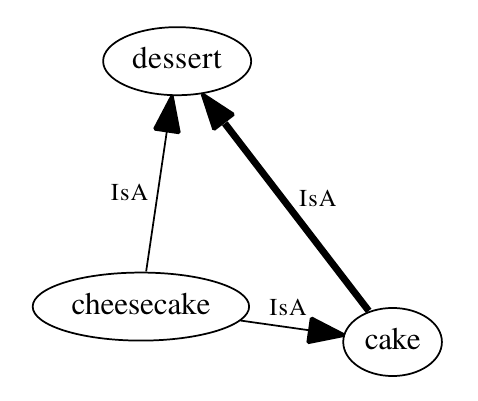}}}
    \hfil
    \subfigure[Mutation type IIb (before)]{\parbox[t]{1.7in}{\centering\includegraphics[width=1.3in]{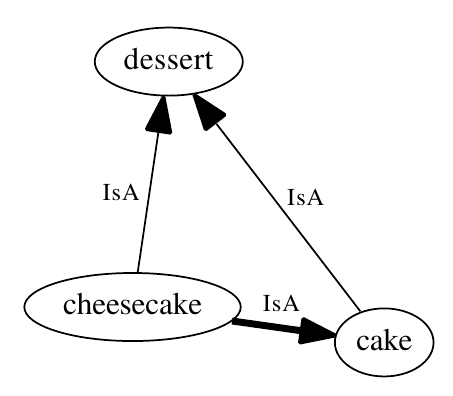}}}
    \hfil
    \subfigure[Mutation type IIb (after)]{\parbox[t]{1.7in}{\centering\includegraphics[width=1.3in]{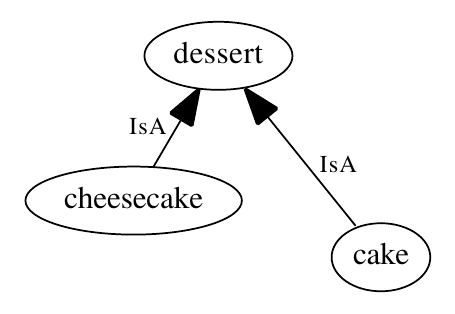}}}
    \hfil
    \subfigure[Mutation type IIIa (before)]{\parbox[t]{1.7in}{\centering\includegraphics[width=1.7in]{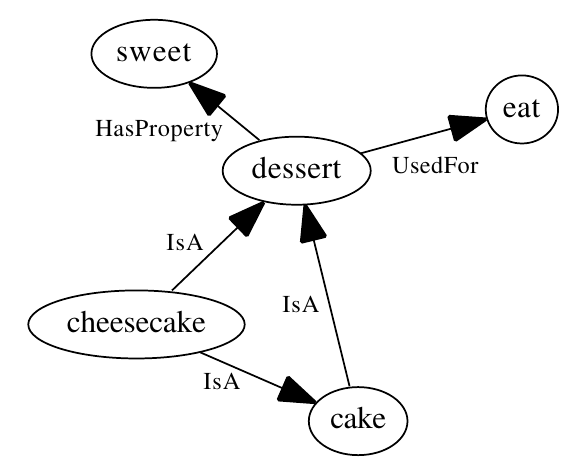}}}
    \hfil
    \subfigure[Mutation type IIIa (after)]{\parbox[t]{1.7in}{\centering\includegraphics[width=1.7in]{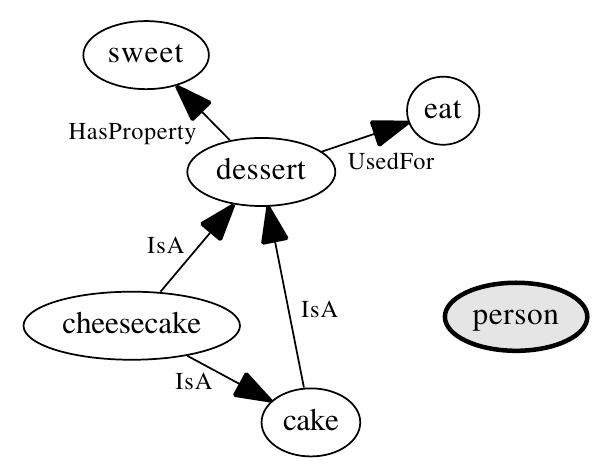}}}
    \hfil
    \subfigure[Mutation type IIIb (before)]{\parbox[t]{1.7in}{\parbox[t]{1.6in}{\centering\includegraphics[width=1.6in]{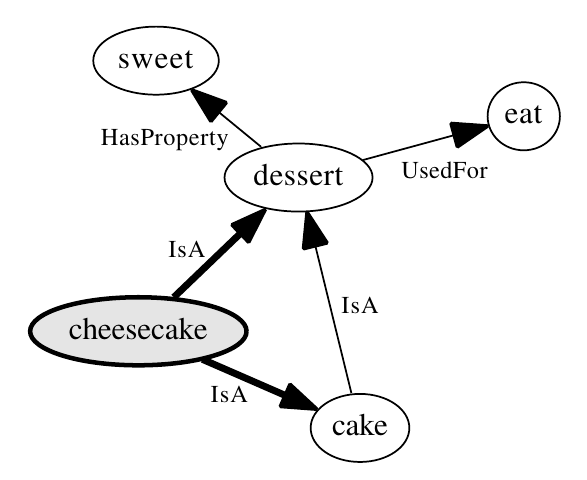}}}}
    \hfil
    \subfigure[Mutation type IIIb (after)]{\parbox[t]{1.7in}{\centering\includegraphics[width=1.7in]{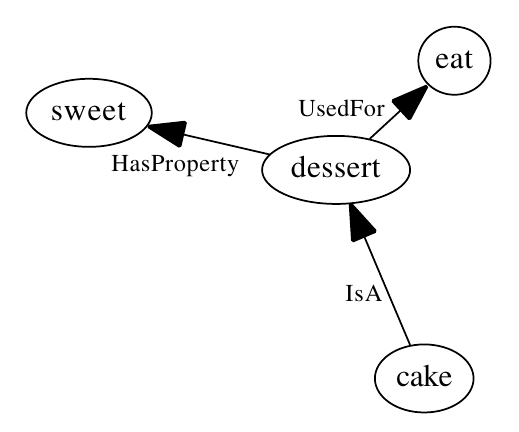}}}
    \hfil
    \subfigure[Mutation type IV (before)]{\parbox[t]{1.7in}{\centering\includegraphics[width=1.7in]{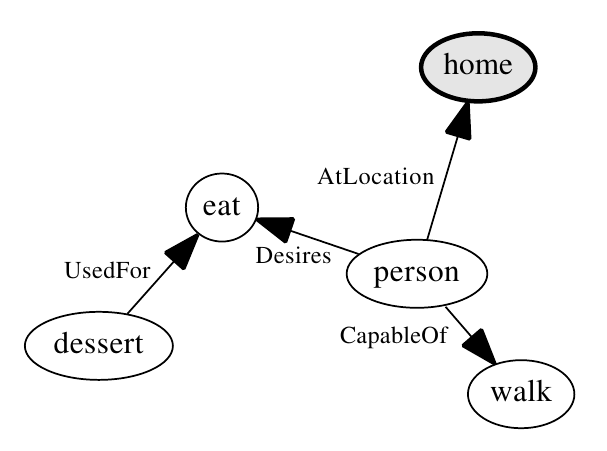}}}
    \hfil
    \subfigure[Mutation type IV (after)]{\parbox[t]{1.7in}{\centering\includegraphics[width=1.7in]{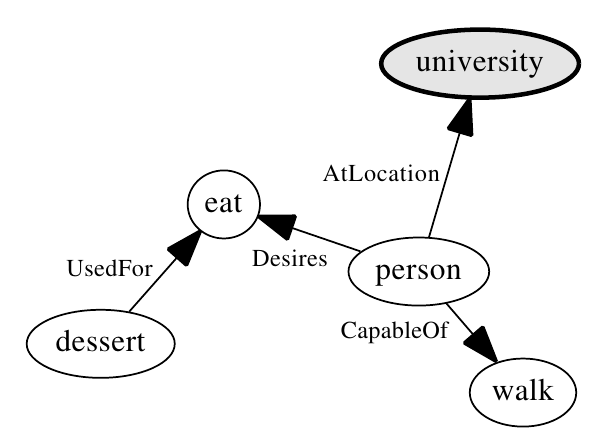}}}
    \caption{Examples illustrating the types of commonsense mutation used in this study.}
    \label{FigureMutation}
  \end{figure*}

  \emph{Type I (concept attachment):} A new concept randomly picked from the set of concepts \emph{attachable} to the parent is attached through a new relation to one of existing concepts (Figure~\ref{FigureMutation} (a) and (b)).

  \emph{Type IIa (relation addition):} A new relation connecting two existing concepts in the parent is added, possibly connecting unconnected clusters within the same network (Figure~\ref{FigureMutation} (c) and (d)).

  \emph{Type IIb (relation deletion):} A randomly picked relation in the parent is deleted, possibly leaving unconnected clusters within the same network (Figure~\ref{FigureMutation} (e) and (f)).

  \emph{Type IIIa (concept addition):} A randomly picked new concept is added to the parent as a new cluster (Figure~\ref{FigureMutation} (g) and (h)).

  \emph{Type IIIb (concept deletion):} A randomly picked concept is deleted with all the relations it is involved in, possibly leaving unconnected clusters within the same network (Figure~\ref{FigureMutation} (i) and (j)).

  \emph{Type IV (concept replacement):} A concept in the parent, randomly picked from the set of those with at least one \emph{interchangeable} concept, is replaced with one of its interchangeable concepts, again randomly picked. Any relations left unsatisfied by the new concept are deleted (Figure~\ref{FigureMutation} (k) and (l)).

  \subsection{Fitness Measure}
  \label{SectionFitness}

  Since the evolving individuals in our approach represent pieces of knowledge, or memes, the fitness measure for evolutionary selection is defined as a function of the represented knowledge. In Section~\ref{SectionExperimentalResults} we define a memetic fitness measure based on the \emph{structure mapping theory} from psychology \cite{Gentner1997}, to evaluate our approach. The simple fitness function used in this introductory study can be extended to take graph-theoretical properties of semantic networks into account, such as the number of nodes or edges, shortest path length, or the clustering coefficient \cite{Steyvers2005}.

  Another interesting possibility is to make the inclusion of certain concepts a requirement, allowing the discovery of memes formed around a given set of seed concepts. This can be also achieved through starting the initialization procedure described in Section~\ref{SectionInitialization} with the given concepts.

  A direct and very interesting application of our approach would be to devise experiments with realistically formed fitness functions modeling selectionist theories of knowledge, which remain untested until this time. One such theory is the \emph{evolutionary epistemology} of Campbell \cite{Bickhard2003}, describing the development of human knowledge and creativity through selectionist principles such as blind variation and selective retention (BSVR).

\section{Experiments and Results}
\label{SectionExperimentalResults}

  To evaluate our approach, we first introduce a fitness measure based on structure mapping. The rest of this section then summarizes our choice of parameters and results from experiments.

  \subsection{Analogy with a Given Semantic Network}

  As a simple and interesting memetic fitness function, we introduce analogical similarity with a given semantic network, utilizing the Structure Mapping Engine (SME) \cite{Falkenhainer1989,Forbus1994}. SME is an algorithm implementing the psychological structure mapping theory of Gentner \cite{Gentner1997}, often cited as the most influential work on modeling analogy-making \cite{French2002}. Using the analogical matching score from SME as a fitness measure, our algorithm can evolve collections of information, or memes, that are analogous to a given one.

  SME is based on the idea that an analogy is a one-to-one mapping from one domain (the base) into another (the target), which correspond, in our fitness measure, to the semantic network supplied at the start and the individual networks whose fitnesses are evaluated by the function. The mapping is guided by the structure of relations between concepts in the two domains, ignoring the semantics of the concepts themselves; and is based on the systematicity principle, where connected knowledge is preferred over independent facts and is assigned a higher matching score. A commonly used example is the analogy between the Solar System and the Rutherford--Bohr model of the atom \cite{Gentner1997}, where $sun$ and $planet$ in the first domain are analogous to $nucleus$ and $electron$ in the second domain. The labels and structure of relations in the two domains (e.g. $\{Attracts(sun, planet)$, $Orbits(planet, sun)$, $\cdots\}$ and $\{Attracts(nucleus$, $electron)$, $Orbits(electron$, $nucleus)$, $\cdots\}$) define and constrain the possible mappings between concepts in the two domains that can be formed by SME.

  We make use of our own implementation of SME based on the original description by Falkenhainer \cite{Falkenhainer1989} and adapt it to the simple concept--relation structure of semantic networks, by mapping the predicate calculus constructs of \emph{entities} into \emph{concepts}, \emph{relations} to \emph{relations}, \emph{attributes} to $IsA$ \emph{relations}, and excluding \emph{functions}.

  \subsection{Results}

  In this introductory study, we adopt values for crossover and mutation probabilities similar to earlier studies in graph-based EA \cite{Pereira1999,Koza2003} (Table~\ref{TableParameters}). We use a crossover probability of $P_c = 0.85$, and a somewhat-above-average mutation rate of $P_m = 0.15$, accounting for the high tendency of mutation postulated in memetic literature\footnote{See Gil-White \cite{GilWhite2008} for a review and discussion of mutation in memetics.}. We employ tournament selection, meaning that for each selection, a ``tournament'' is held among a few randomly chosen individuals, and the more fit individual of each successive pair is the winner according to a winning probability. In our experiments, we subject a population of $Pop_{size} = 200$ individuals to tournament selection with tournament size $S_{size} = 8$ and winning probability $S_{prob} = 0.8$.

\begin{figure}[!t]
    \centering
    \subfigure[Given semantic network, 10 concepts, 11 relations (base domain)]{\parbox[t]{3.4in}{\centering\includegraphics[width=2.8in]{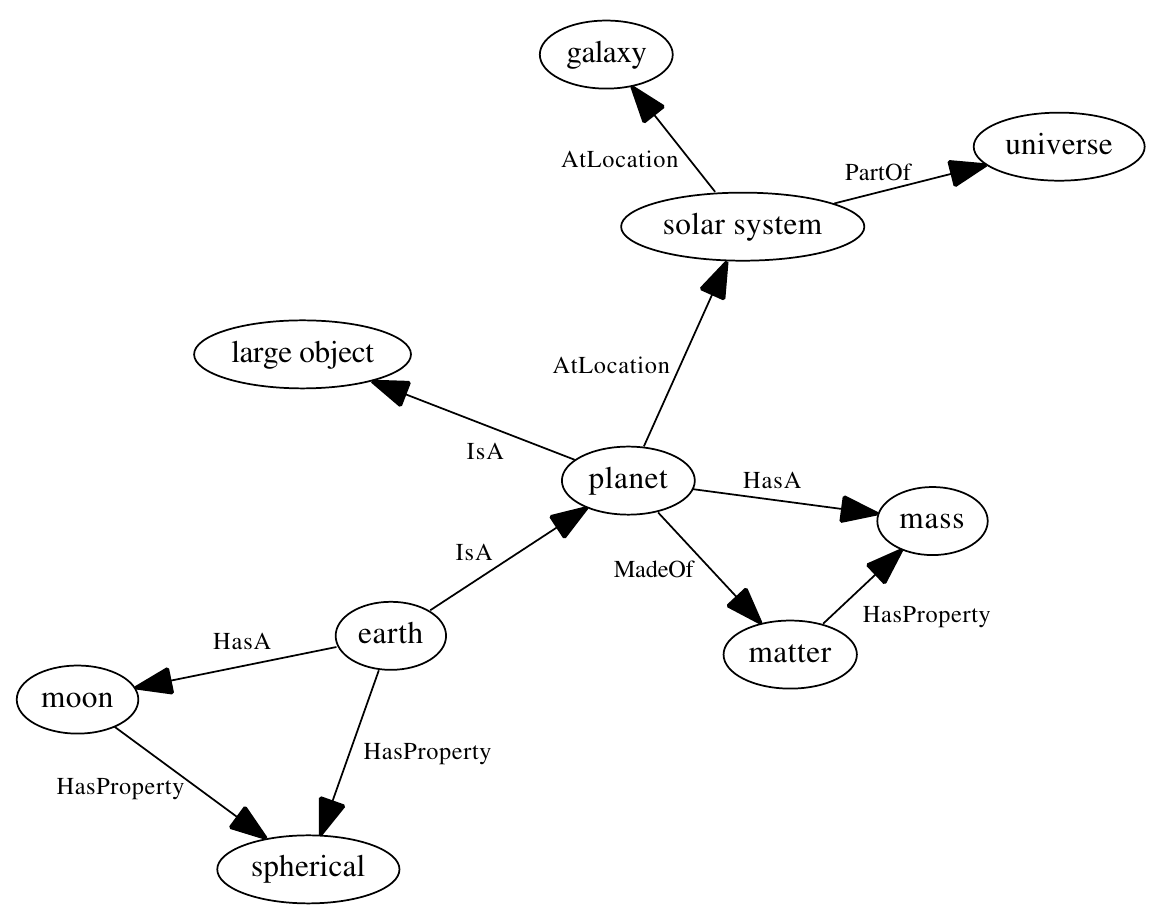}}}
    \hfil
    \subfigure[Evolved individual, 9 concepts, 9 relations (target domain)]{\parbox[t]{3.4in}{\centering\includegraphics[width=2.4in]{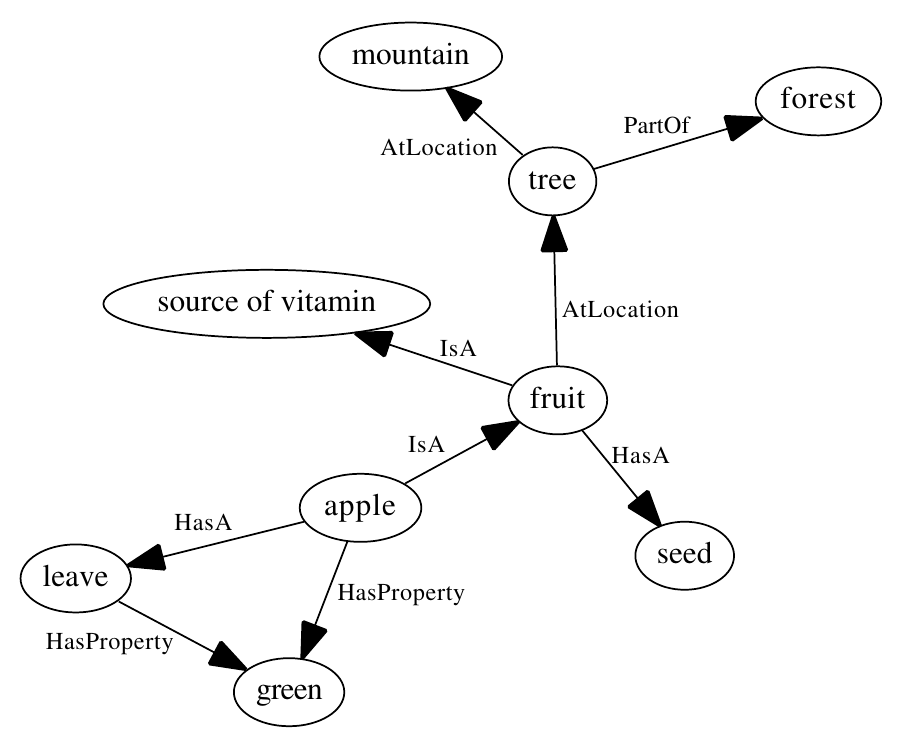}}}
    \caption{Experiment 1: The evolved individual is encountered after 35 generations, with fitness value 2.8. Concepts and relations of the individual not involved in the analogy are not shown here for clarity.}
    \label{FigureExperiment1}
  \end{figure}

\begin{figure}[!t]
    \centering
    \subfigure[Given semantic network, 11 concepts, 11 relations (base domain)]{\parbox[t]{3.4in}{\centering\includegraphics[width=2.5in]{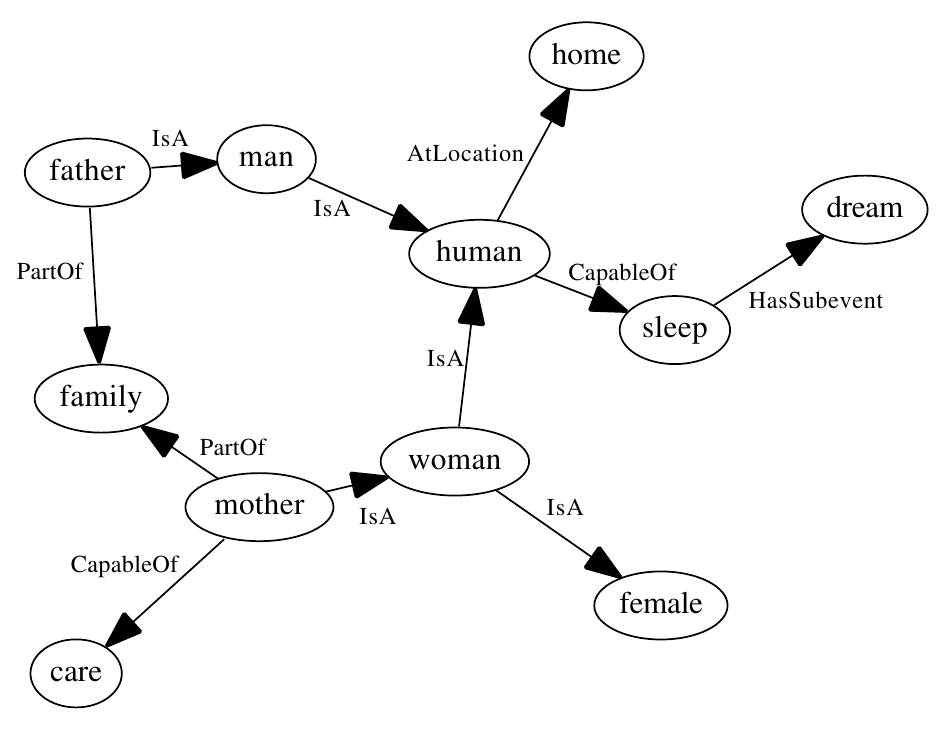}}}
    \hfil
    \subfigure[Evolved individual, 10 concepts, 9 relations (target domain)]{\parbox[t]{3.4in}{\centering\includegraphics[width=2.7in]{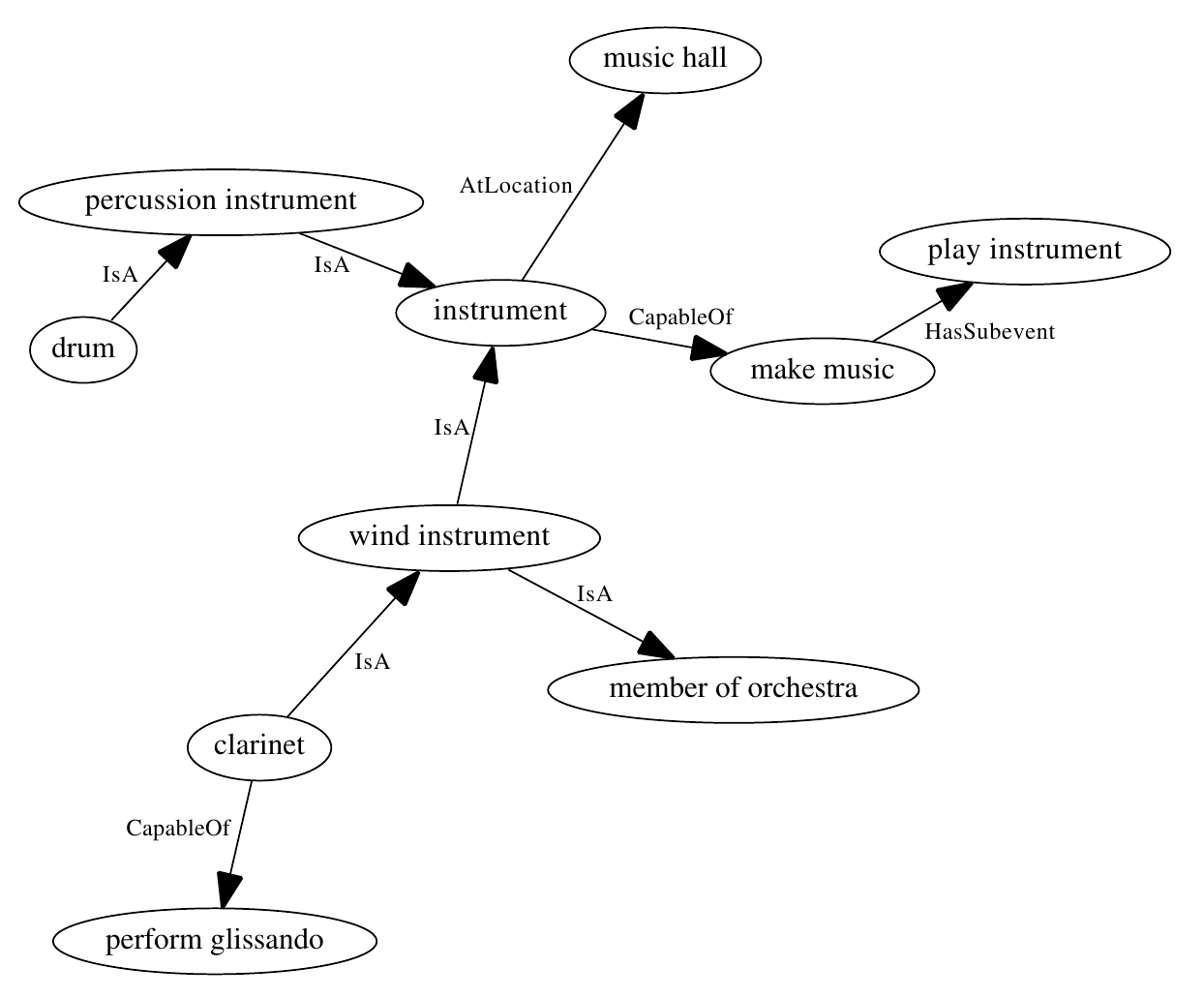}}}
    \caption{Experiment 2: The evolved individual is encountered after 42 generations, with fitness value 2.7. Concepts and relations of the individual not involved in the analogy are not shown here for clarity.}
    \label{FigureExperiment2}
  \end{figure}

\begin{figure}[!t]
    \centering
    \subfigure[]{\includegraphics[width=2.9in]{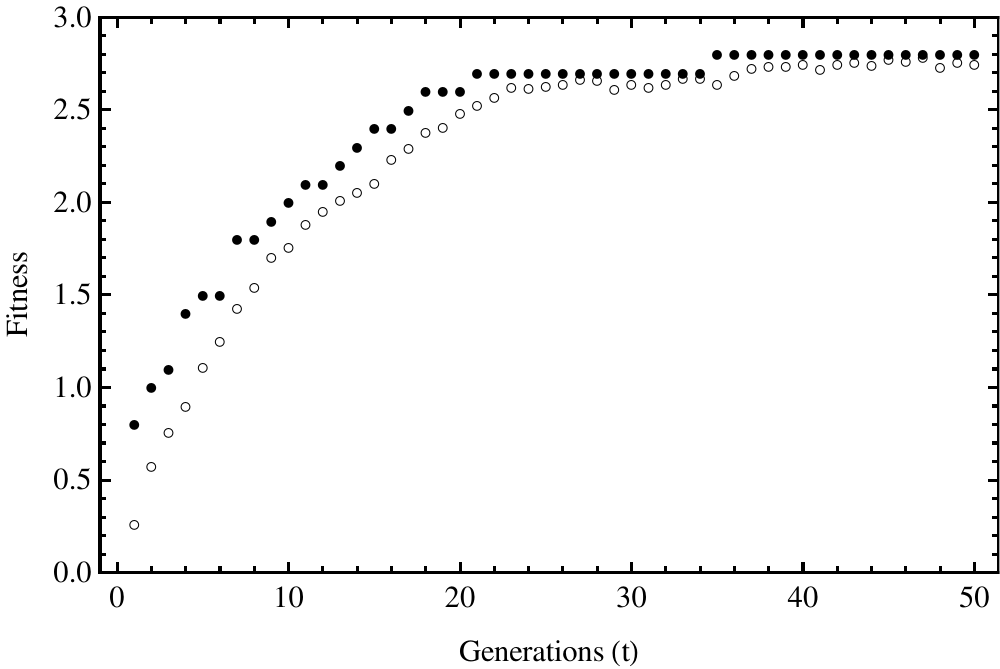}}
    \subfigure[]{\includegraphics[width=2.9in]{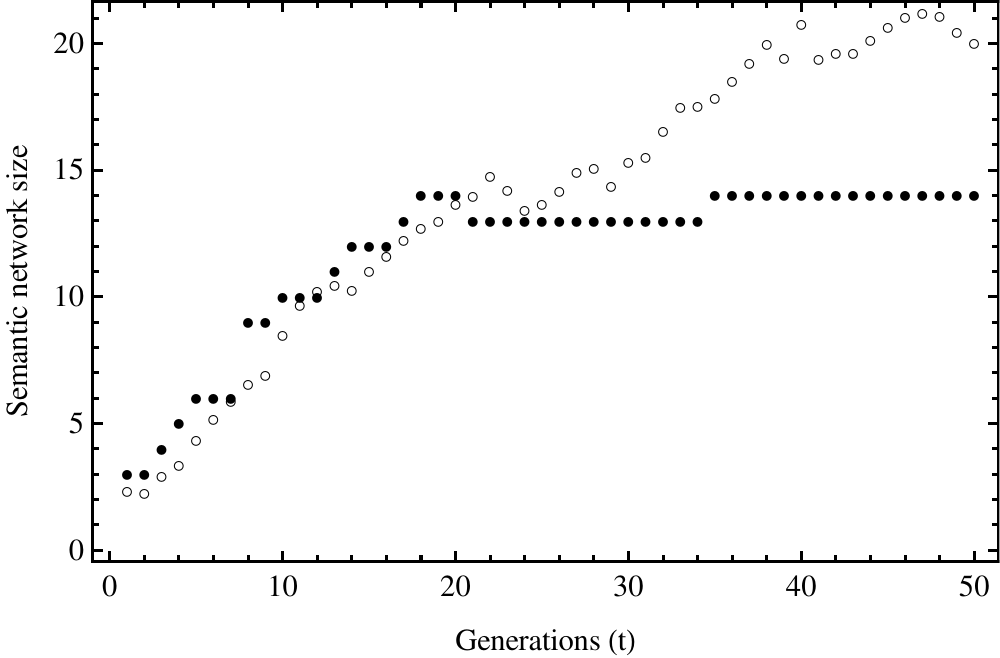}}
    \caption{Evolution of (a) fitness and (b) semantic network size during the course of an experiment with parameters given in Table~\ref{TableParameters}. Filled circles represent the best individual in a generation, while the empty circles represent population average. Network size is taken to be the number of relations (edges) in the semantic network.}
    \label{FigurePlots}
  \end{figure}

  Using this parameter set, we present the results from two runs of experiment: evolved analogies for a network describing some basic astronomical knowledge are shown in Figure~\ref{FigureExperiment1} and for a network of familial relations in Figure~\ref{FigureExperiment2}. We show in Figure~\ref{FigurePlots} (a) the progress of the best and average fitness in the population during the run that produced the results in Figure~\ref{FigureExperiment1}. The best and average size of semantic networks forming the individuals are shown in Figure~\ref{FigurePlots} (b). We observe that evolution asymptotically reaches a fitness plateau after about 40 generations. This coincides roughly with the point where the size of the best individual becomes comparable with that of the given base semantic network (Figure~\ref{FigureExperiment1}), after which improvements in the one-to-one analogy become sparser. Our experiments demonstrate that the proposed algorithm is capable of spontaneously creating collections of knowledge analogous to the one given in a base semantic network, with very good performance. In most cases, our implementation was able to reach extensive analogies within 50 generations and reasonable computational time.

  \begin{table}[!t]
    \renewcommand{\arraystretch}{1.3}
    \caption{Parameters used during experiments}
    \label{TableParameters}
    \centering
    \begin{tabular}{lcc}
    \FL
     & Parameter & Value
    \ML
    Evolution & Population size ($Pop_{size}$) & 200
    \NN
     & Crossover probability ($P_c$) & 0.85
    \NN
     & Mutation probability ($P_m$) & 0.15
    \ML
    Semantic networks & Max. initial concepts ($C_{max}$) & 5
    \NN
     & Min. relation score ($R_{min}$) & 2.0
    \NN
     & Timeout ($T$) & 10
    \ML
    Selection & Type & Tournament
    \NN
     & Tournament size ($S_{size}$) & 8
    \NN
     & Tournament win prob. ($S_{prob}$) & 0.8
    \NN
     & Elitism & Employed
    \LL
    \end{tabular}
  \end{table}

\section{Conclusion and Future Work}
  \label{SectionConclusion}
  In summary, we have presented a novel evolutionary algorithm that employs semantic networks as evolving individuals, paralleling the evolutionary model of cultural variation and selection in the existing field of memetics. This algorithm, to our knowledge, is the first of its kind. The use of semantic networks provides a suitable basis to implement the representation and manipulation of memes---in the sense of units of culture, or knowledge, as it was put forth by Dawkins \cite{Dawkins1989}. We have introduced preliminary versions of variation operators that work on this representation, utilizing knowledge from commonsense knowledge bases. We have also contributed a memetic fitness measure based on the structure mapping theory of Gentner \cite{Gentner1997} with a firm basis in psychology, to evaluate the feasibility and performance of our work. Considering other possible fitness measures, we hope that our approach can serve as a computational tool for implementing and experimenting with different theories of cultural evolution encountered in memetics, such as evolutionary epistemology and cultural selection theory. The facts that:

  \begin{itemize}
    \item the algorithm is exclusively on the evolution of knowledge itself under a fitness measure defined as a function of the represented knowledge;
    \item and that it uses an encoding based on knowledge representation (in the form of semantic networks), utilizing memetic operators specifically created for the modification of these
  \end{itemize}

  set it apart from existing MA, where the concept of memetics has been used as a synonym for an hybridization of local refinement by individual learning after a classical global EA sampling. 

  The proposed algorithm is highly relevant from the perspective of computational creativity \cite{Wiggins2006,Duch2007}, especially for tasks such as conceptual blending \cite{Fauconnier2009} and story generation \cite{Riedl2010}. We believe that our approach can fit into both classifications of creativity discussed by Boden \cite{Boden2009}: (1) \emph{exploratory creativity}, where the commonsense nature of our memetic operators addresses the criticism about the lack of world knowledge\footnote{Quting Boden \cite{Boden2009}: \emph{``What's missing, as compared with the human mind, is the rich store of world knowledge (including cultural knowledge) that's often involved.''}}; and (2) \emph{transformational creativity}, where the potential of evolutionary approaches have already been noted. Within the field of analogical reasoning, virtually all of existing research has been focused on the discovery and assessment of possible analogical mappings between two given domains \cite{French2002}, while our approach provides a novel technique for the open-ended discovery of information analogous to a given domain, together with the analogical mapping.

  For future work following this study, we acknowledge several lines of research concerning the design of the algorithm and its possible applications. Regarding the design, the research would benefit from exploring different types of mutation and crossover \cite{Pereira1999,Katagiri2002}. It would be significant to ground the design of such operators on existing theories of cultural transmission and modification, discussed in sociological theories of knowledge.

  Regarding the applications, in an upcoming paper we intend to focus on different types of computational creativity that would be achievable with this model. In the long term, we argue that if a theoretically sound basis for memetic variation and inheritance can be put in place, and together with realistic memetic fitness measures, our approach can enable ``computational memetic simulations'' analogous to those in computational biology, such as genetic drift or coalescent theories.

\section*{Acknowledgment}
  \addcontentsline{toc}{section}{Acknowledgment}
  This work was supported by a JAE-Predoc fellowship from CSIC, and the research grants: 2009-SGR-1434 from the Generalitat de Catalunya, CSD2007-0022 from MICINN, and Next-CBR TIN2009-13692-C03-01 from MICINN.


\bibliographystyle{IEEEtran}
\bibliography{BaydinEvolution}

\end{document}